\documentclass{article}
\usepackage[utf8]{inputenc}

\usepackage[lofdepth,lotdepth,caption=false]{subfig}
\usepackage{fancyhdr}
\usepackage{hyperref}
\usepackage[tbtags]{amsmath}
\usepackage{amsmath, amssymb, graphicx}
\usepackage{xspace}
\usepackage{braket}
\usepackage{color}
\usepackage{setspace}
\usepackage{comment}
\usepackage{outlines}
\usepackage{xcolor,colortbl}
\usepackage{cleveref}
\usepackage{changepage}
\usepackage{lscape}
\usepackage{geometry}
\usepackage{pdflscape}
\usepackage{bm}
\usepackage{caption}
\usepackage{float}

\DeclareMathOperator*{\argmax}{arg\,max}

\title{Discriminative classification with generative features: bridging Naive Bayes and logistic regression}

\author{
  Zachary Terner\thanks{Corresponding author: zterner@ucsb.edu. 
  This author was partly funded by the National Science Foundation Graduate Research Fellowship.} \and
  Alexander Petersen\thanks{Brigham Young University} \and
  Yuedong Wang\thanks{University of California, Santa Barbara}
}
\date{November 2025}

\usepackage{natbib}
\usepackage{graphicx}

\begin{document}

\maketitle

\begin{abstract}
    We introduce Smart Bayes, a new classification framework that bridges generative and discriminative modeling by integrating likelihood–ratio–based generative features into a logistic-regression-style discriminative classifier. From the generative perspective, Smart Bayes relaxes the fixed unit weights of Naive Bayes by allowing data-driven coefficients on density-ratio features. From a discriminative perspective, it constructs transformed inputs as marginal log-density ratios that explicitly quantify how much more likely each feature value is under one class than another, thereby providing predictors with stronger class separation than the raw covariates. To support this framework, we develop a spline-based estimator for univariate log-density ratios that is flexible, robust, and computationally efficient.  Through extensive simulations and real-data studies, Smart Bayes often outperforms both logistic regression and Naive Bayes. Our results highlight the potential of hybrid approaches that exploit generative structure to enhance discriminative performance.
\end{abstract}

\section{Introduction}
The challenge of classifying new observations into one of two (or more) classes is an old and ongoing problem in statistics and machine learning. Many existing classifiers can be categorized as being either \emph{generative} or \emph{discriminative}. As noted by Ng and Jordan~\cite{ng2002discriminative}, generative classifiers learn a model of the joint probability $p(x,y),$ where $x$ represents the inputs and $y$ the response, and use Bayes' rule to compute $p(y|x)$ to make predictions. Discriminative classifiers, however, model $p(y|x)$ explicitly without specifying the marginal distribution of $x$. Since this approach is more direct, these classifiers often perform better in terms of minimizing misclassification rates. Using the generative-discriminative pair of Naive Bayes and logistic regression, Ng and Jordan suggest that discriminative classifiers tend to have lower asymptotic errors than their generative counterparts, but a generative classifier may approach its asymptotic error rate more quickly. Thus, Naive Bayes may outperform logistic regression in settings of small sample sizes, but logistic regression should perform better as sample sizes increase.

By virtue of this pairing, Naive Bayes and logistic regression are fundamentally connected. The Naive Bayes classifier, rather than estimate the joint density of each class, invokes the factorization theorem to factor the joint densities, conditional on class membership, into the products of assumed independent marginals. Taking the log of this product allows one to compute the log marginal density ratio for each feature. New observations are assigned to a class if the sum of the log density ratios exceeds or falls below a threshold dependent on the prior class probabilities. In contrast, logistic regression avoids computing density ratios and learns a model of $p(y|x)$ using the features themselves. Building a model with logistic regression can also provide interpretations that one does not normally gain from Naive Bayes. 



In this paper, we demonstrate that incorporating generative features into a discriminative classification framework can outperform both standard Naive Bayes and conventional discriminative methods. Specifically, our simulations and empirical experiments demonstrate that a logistic regression with marginal log density ratios as features can outperform not only Naive Bayes but also a logistic regression using the features themselves. 

To build this classifier, we compute marginal log-density ratios as done in Naive Bayes. Rather than applying the Naive Bayes rule directly, these ratios are entered into a logistic regression model as features. Under this formulation, Naive Bayes arises as a special case in which all model coefficients are fixed to one. We refer to the proposed method as Smart Bayes (SB), reflecting its more flexible utilization of generative information. 
Importantly, the weights of the log density ratios can be interpreted similarly to the coefficients in a logistic regression. This approach can also be applied to other classification methods, such as decision trees and support vector machines, which may similarly benefit from incorporating log-density ratios as features.

Smart Bayes can also be interpreted through the lens of the likelihood ratio. In Smart Bayes, the original features in a logistic regression are replaced with log-transformed marginal density ratios. These may be more discriminative than the raw inputs, since they quantify how likely it is that each component of the observed feature vector originated from one class over the other. 
Therefore, they can serve as more informative predictors for classification than the original features themselves.

To treat density ratios as features in a regression, one must first construct estimates of density ratios. The estimation of density ratios pervades many areas of statistics, from building classification models, outlier detection~\cite{hido2011statistical}, changepoint detection~\cite{liu2013change}, deep learning~\cite{uehara2016generative}, and machine learning in general~\cite{sugiyama2012density}. See~\cite{sugiyama2011direct} for additional application areas and references regarding density ratio estimation. We introduce a new method of estimating density ratios directly using splines, which is based on previous techniques by Silverman and Kanamori~\cite{silverman1978density, kanamori2009least}. Although we use this method to estimate marginal log density ratios, it can be applied more generally to estimate joint log density ratios of features.

The paper is structured as follows. In Section~\ref{sect-lrnb}, we review Naive Bayes, logistic regression, and their relationship. In Section~\ref{sect-dcgf}, we introduce Smart Bayes and describe some of its properties. 
Section~\ref{sect-density} introduces the method of direct estimation of marginal density ratios using splines, which we use in the simulation and real data examples in Sections~\ref{sect-expt}. These real data examples are constructed using the same datasets as Ng and Jordan~\cite{ng2002discriminative} to show Smart Bayes's behavior relative to Naive Bayes and logistic regression. Empirically, Smart Bayes outperforms both alternatives on a variety of datasets. Section~\ref{sect-discussion} concludes the paper with a discussion.


\section{Logistic regression and Naive Bayes}
\label{sect-lrnb}
\label{sect-bayesclassifiers}

\subsection{Naive Bayes}
The Bayes classifier is known to minimize misclassification rates~\cite{bickel2004some,young1974classification,domingos1997optimality, james2013introduction}. Let $Y$ be our class labels and denote $\mathbf{X}$ as our $p$ multivariate predictors. The Bayes classifier can be written as

\begin{equation}
\label{genbayes}
    \hat{y} = \argmax_y P(Y = y | \mathbf{X} = \mathbf{x}).
\end{equation}
In other words, we assign new observations $x$ to the class which has the maximum posterior probability~\cite{fox}. In the two-class problem, where $Y \in \{0, 1\},$ let
\begin{equation}
\label{twobayes}
    f(\mathbf{x}) = \frac{P(Y = 1 | \mathbf{X} = \mathbf{x})}{P(Y = 0 | \mathbf{X} = \mathbf{x})}.
\end{equation}
The Bayes classifier assigns observations to class $Y = 1$ if $f(\mathbf{x}) > 1.$ In this case, we compute the posterior conditional probabilities of the two classes and take the ratio to determine which class has the higher posterior probability. By applying Bayes' theorem,

\begin{equation}
\label{twobayes2}
f(\mathbf{x}) = \frac{P(Y = 1) P(\mathbf{X} = \mathbf{x} | Y = 1) }{ P(Y = 0) P(\mathbf{X} = \mathbf{x} | Y = 0)},
\end{equation}
where we have switched the conditioning to require a density of the data conditional on the class, $P(\mathbf{X} = \mathbf{x}|Y = y).$ The class probabilities, $P(Y = y),$ can be estimated by computing the proportion of instances in each class. However, the conditional probabilities in~\eqref{twobayes2} can be very difficult to estimate in practice, especially when $p$ is large, due to the curse of dimensionality~\cite{shalizi_2016}.

One common assumption used to compute the joint densities in~\eqref{twobayes2} is to assume that the predictor variables are independent, conditional on class membership. Let $g_y(\mathbf{x})$ be the density function of $\mathbf{X}$ conditional on $Y=y$. Let $\mathbf{X}=(X_1,\ldots,X_p)$ and $g_{yk}$ be the marginal density function of $X_k$ conditional on $Y=y$. By the factorization theorem, if the predictor variables are independent conditional on the class label, we can write
\begin{equation}
\label{twobayes3}
f(\mathbf{x}) = \frac{P(Y = 1) g_1(\mathbf{x}) }{ P(Y = 0) g_0(\mathbf{x})} = \frac{P(Y = 1) \prod_{k=1}^p g_{1k}(x_k) }{ P(Y = 0) \prod_{k=1}^p g_{0k}(x_k) }.
\end{equation}

Since~\eqref{twobayes3} contains only products, we can take the log to convert the products into a sum. This creates the traditional Naive Bayes classifier, which we denote as $f_n(x):$
\begin{equation}
\label{Naivebayes}
f_n(\mathbf{x}) = \log \frac{P(Y = 1)}{P(Y = 0)} + \sum_{k = 1}^p \log \frac{ g_{1k}(x_k)}{g_{0k}(x_k)}.
\end{equation}
In~\eqref{Naivebayes}, we simply need to compute the log density ratio of each marginal variable. The class label is assigned based on whether the sum of the log density ratios and the prior probabilities is positive ($Y = 1$) or negative ($Y = 0$).

\subsection{Logistic regression}
Working on the log density ratio
\begin{equation}
\label{logistic}
    f_l(\mathbf{x}) \triangleq \log \frac{P(Y = 1 | \mathbf{X} = \mathbf{x})}{P(Y = 0 | \mathbf{X} = \mathbf{x})}
\end{equation}
directly can be justified by the exponential criterion~\cite{friedman2000additive}. A simple model for $f_l$ is the logistic regression model
\begin{equation}
\label{logistic-mod}
    f_l(\mathbf{x}) = \beta_0 + \sum_{k=1}^p \beta_k x_k.
\end{equation}
Instead of computing marginal density ratios, logistic regression uses the features $x_k$ directly to build the function that separates the classes. The coefficients $\beta_k$ and an intercept $\beta_0$ can be learned as maximizers of the likelihood~\cite{givens2012computational}. 
New observations are assigned to Class 1 if the estimate of $f_l(\mathbf{x})$ is above a certain threshold, and to Class 0 otherwise.

Logistic regression is considered discriminative since it directly connects the features $x_k$ to $P(Y|\mathbf{X} = \mathbf{x})$. In contrast, Naive Bayes is generative since it approximates the joint probability $P(\mathbf{X}, Y)$ and uses Bayes rule to calculate $P(Y |\mathbf{X} = \mathbf{x}).$ Ng and Jordan show that generative classifiers approach their asymptotic error rate more quickly than discriminative classifiers, though discriminative classifiers have a lower asymptotic error rate~\cite{ng2002discriminative}. Therefore, Naive Bayes is expected to outperform logistic regression only when sample sizes are small.

Both Naive Bayes and logistic regression methods have their limitations. To approximate the joint probability, Naive Bayes ignores any dependence between the features by (naively) assuming the features are independent. The conditional independence between predictors is a strong assumption that, in general, does not hold in practice. 
Logistic regression does not make this assumption, but does assume a parametric model for the log odds. 

\section{Smart Bayes}
\label{sect-dcgf}
\label{sect-smartbayes}

Section \ref{sect-SB} introduces the Smart Bayes method, and Section \ref{sect-density} outlines our procedure for estimating the marginal log density ratio. We discuss how to interpret the coefficients in Smart Bayes in Section~\ref{sect-interpretation}.


\subsection{Discriminative classifier with generative features}
\label{sect-SB}

When the class-conditional distributions of $\mathbf{x}$ are specified, the logistic regression classifier can be written as
\begin{equation}
\label{dcgf2}
    f_l(\mathbf{x}) = \log\frac{P(Y = 1)}{P(Y = 0)} + \log \frac{g_1(\mathbf{x})}{g_0(\mathbf{x})}.
\end{equation}
From a generative perspective, logistic regression attempts to approximate the log-density ratio
using a linear combination of the original features $x_k$ and an intercept. The key idea of our proposed method is that an affine function of the marginal log-density ratios may provide a more accurate approximation to
$f_l(\mathbf{x}).$ 

Define generative features $z_k = \log\{ g_{1k}(x_k) / g_{0k}(x_k) \}.$ When the features $X_k$ are mutually conditionally independent, we can write
\begin{equation}
\label{dcgf3}
    f_l(\mathbf{x}) =
    \log\frac{P(Y = 1)}{P(Y = 0)} + 
    \sum_{k=1}^p z_k,
\end{equation}
which coincides with the Naive Bayes classifier $f_n$ in~\ref{Naivebayes}.

To accommodate dependence among the features, the proposed Smart Bayes approach, defined as $f_s$ below, fits the following logistic regression model with generative features $z_k$:
\begin{equation}
    \label{dcgf4}
    f_s(\mathbf{x}) = \alpha_0 + \sum_{k=1}^p \alpha_k z_k.
\end{equation}
The Smart Bayes method extends the Naive Bayes classifier by replacing its fixed unit coefficients with learnable weights $\alpha_k$, allowing the contribution of each generative feature to be determined from the data. At the same time, Smart Bayes can be viewed as an extension of logistic regression, obtained by substituting the original features $x_k$ with possibly more informative generative features $z_k$. Thus,  Smart Bayes retains the strengths of both approaches by leveraging generative information while preserving the flexibility of discriminative modeling.

As in standard logistic regression, interaction terms or higher-order transformations of $z_k$ may be incorporated to capture more complex relationships. Moreover, the generative features $z_k$ can be used as predictors in other machine-learning classifiers, such as support vector machines or decision trees, potentially improving performance by leveraging informative log-density-ratio features.

The Smart Bayes classifier offers a few notable properties. When both classes are Gaussian, Smart Bayes can recover the optimal Bayes classifier in two cases: when the two classes have a difference in means but a shared covariance, and when the two classes have different means and different covariances, but the same variances. In general, Smart Bayes can recover the optimal Bayes solution when the marginal density ratios $z_k$ are all linear in $\mathbf{x}$. It is additionally easy to see that there are many cases where coefficients in the Smart Bayes classifier should not only differ from one, but should in fact be negative. See~\cite{terner2020classification} for additional details.

\subsection{Direct density ratio estimation using splines}
\label{sect-density}
In this section, we introduce a method for estimating marginal density ratios, which are typically unknown in practice. We extend Silverman’s method~\cite{silverman1978density} to a more general model space and a broader class of penalties. A related approach appears in~\cite{kanamori2009least}, but the method presented here is more general.

We estimate the conditional density ratio for each random variable in $\mathbf{X}=(X_1,\ldots,X_p)$. For simplicity, we omit the subscript in the following discussion. 
Suppose that we observe $(X,Y)$ where $Y = 1$ or $Y = 0,$ $X \in \mathcal{X}$ and $\mathcal{X}$ is an arbitrary set. Let $g_0(x)$ and $g_1(x)$ denote the conditional density of $X | Y = 0$ and $X | Y = 1$, respectively.

Using Bayes' theorem, we can write 
\begin{equation}
    p(x) \triangleq P(Y = 1|X = x) = \frac{\pi_1 g_1(x)}{\pi_1 g_1(x) + \pi_0 g_0(x)} = \frac{r d(x)}{1 + r d(x)},
\end{equation}
where $r$ represents the ratio of prior probabilities, $\pi_1/\pi_0,$ and $d(x)$ represents the density ratio, $g_1(x) / g_0(x).$

The logistic probability can be written as 
\begin{equation}
    \eta(x) = \log \frac{p(x)}{1 - p(x)} = \log r + \log d(x).
\end{equation}

Given data $\{(x_i, y_i),~i=1,\ldots, n \}$, we estimate $\eta(x)$ nonparametrically as the minimizer of the penalized likelihood: 
\begin{equation}
\label{penlik}
    -\sum_{i=1}^n y_i \eta(x_i) + \sum_{i=1}^n \log (1 + \exp{\eta(x_i)}) + \lambda J(\eta) ,
\end{equation}
where $\eta$ belongs to a reproducing kernel Hilbert space (RKHS) ${\cal H}$ defined on the domain ${\cal X}$, $J(\eta)$ is a penalty, and $\lambda$ is a smoothing parameter~\cite{wahba1995smoothing,wang2011smoothing}. The choices of the model space ${\cal X}$ and penalty $J(\eta)$ depend on the domain ${\cal X}$ and prior knowledge.  
The assumption that $X$ takes values in an arbitrary set ${\cal X}$ allows the framework to accommodate different types of random variables in a unified manner. For example, if $X$
is a univariate continuous variable with bounded support, we may take ${\cal X}$ as the support 
 and choose ${\cal H}$ to be an appropriate Sobolev space; if $X$ is supported on $\mathbb{R}$, a thin-plate spline space may be used instead. See~\cite{wang2011smoothing} for additional examples of suitable model spaces. More generally, when prior knowledge indicates that a subset of variables in $\mathbf{X}$ is correlated, one may estimate their joint density rather than marginal densities for individual variables. Nevertheless, in the remainder of this work, we focus on methods based solely on estimates of marginal densities.
 
Standard smoothing methods can be used to obtain the minimizer of the penalized likelihood with $y_i$ as binary outcomes. The smoothing parameter $\lambda$ may be 
selected using generalized cross-validation (GCV), generalized maximum likelihood (GML), or other established criteria~\cite{wang2011smoothing}. The estimate of the log density ratio using the RKHS, $\hat{d},$ is defined as
\begin{equation}
\label{dheq}
    \hat{d}(x) \triangleq \hat{\eta}(x) - \log \hat{r},
\end{equation}
where $\hat{r}$ is an estimate of $r$. A simple estimate is $\hat{r}=\sum_{i=1}^n y_i/ (n - \sum_{i=1}^n y_i).$ In comparison, Silverman~\cite{silverman1978density} justified his method via the connection between inhomogeneous Poisson processes and density functions. He used the likelihood conditioned on sufficient statistics and treated the sum of intensity functions as a nuisance parameter. Therefore, his method is limited to density ratios defined on the real line. Note that the negative log-likelihood in the first two terms in~\eqref{penlik} has the same form as the conditional log-likelihood presented in~\cite{silverman1978density}.  Therefore, with the same model space and penalty, these two approaches lead to the same estimate.


The Bayes rule based on this estimated density ratio is equivalent to nonparametric logistic regression: the Bayes rule assigns an observation to Class 1 when $rd(x) \geq 1$, which is equivalent to $\eta(x) \geq 0$~\cite{wahba1995smoothing}.
In Smart Bayes, we estimate marginal density ratios and then fit logistic regression models with these estimated log density ratios as features. Specifically, for multivariate $\mathbf{X} = (X_1, \dots, X_p)$, we first estimate marginal density ratios $d_k(x_k)=g_{1k}(x_k)/g_{0k}(x_k)$ for $k=1,\ldots,p$ and then fit a logistic regression model with $z_k=\log d_k(x_k)$ as features.  Therefore, we only need to solve~\eqref{penlik} for univariate spline models for each marginal density ratio. Note that Bayes rule traditionally uses the full likelihood $f(y|x) f(x)$ while logistic regression uses the conditional likelihood, $f(y|x).$ In our approach, we estimate the density ratio using the conditional likelihood and treat the prior density $f(x)$ as a nuisance parameter.

\subsection{Interpretation of Smart Bayes coefficients}
\label{sect-interpretation}
As with standard logistic regression models, $\exp(\alpha_k)$ in the proposed Smart Bayes model represents the hazard ratio associated with a one-unit increase in the log likelihood ratio $z_k$.  In Smart Bayes, a $t\%$ change in $z_k$ corresponds to a factor $(1 + t/100)$ change in the likelihood or density ratio. 

Assume $X \mid Y = y$ follows an exponential family distribution,
\[
f(x \mid y) = \exp\left\{ \frac{\theta(y)x - b(\theta(y))}{a(\phi)} + c(x, \phi) \right\},
\]
where $\theta$ and $\phi$ denote the canonical and dispersion parameters. Under this model, the log density ratio $z$ (omitting the subscript) is a linear transformation of $x$:
\[
z(x) = \frac{\theta(1) - \theta(0)}{a(\phi)}\, x - \big( b(\theta(1)) - b(\theta(0)) \big) 
      = \gamma x + \delta,
\]
where $\gamma = \{\theta(1)-\theta(0)\}/a(\phi)$ and $\delta = b(\theta(0)) - b(\theta(1))$.
The resulting scaling factor $\gamma$ has the following forms for several common distributions:
\[
\gamma = 
\begin{cases}
\dfrac{\mu_1 - \mu_0}{\sigma^2}, & \text{Gaussian}, \\[2mm]
\operatorname{logit}(\mu_1) - \operatorname{logit}(\mu_0), & \text{Bernoulli}, \\[2mm]
\log \mu_1 - \log \mu_0, & \text{Poisson},
\end{cases}
\]
where $\mu_y = \mathrm{E}(X \mid Y=y)$ and $\sigma^2 = \mathrm{Var}(X \mid Y=y)$.
In these cases, a one-unit increase in $z$ corresponds to an increase in $x$ of size $1/\gamma$, the reciprocal of the separation between the two classes.

\section{Numerical Experiments}
\label{sect-expt}
Below we evaluate the performance of Smart Bayes versus Naive Bayes and logistic regression on simulated and empirical data. We begin by showing the performance of Smart Bayes on data generated from the Gaussian and t-distributions. We then present the performance of these classifiers on seven empirical datasets. 

All logistic regressions plotted below were fit using the standard $\texttt{glm}$ function in R. The naive Bayes classifiers were built using the $\texttt{naive\_bayes}$ function from the $\texttt{naivebayes}$ package in R~\cite{naivebayes}. To estimate the marginal log density ratios, we used the \texttt{gam} function with \texttt{family = "binomial"} from the \texttt{mgcv} package~\cite{gambook} in R. The code used to generate the results in this paper is available online at \href{https://www.github.com/zt8zf/smartbayes}{GitHub}.

\subsection{Multivariate Gaussian and t-distribution simulations}
\label{sect-subsim}
\begin{figure}[H]
\centering
\includegraphics[width=.9\textwidth]{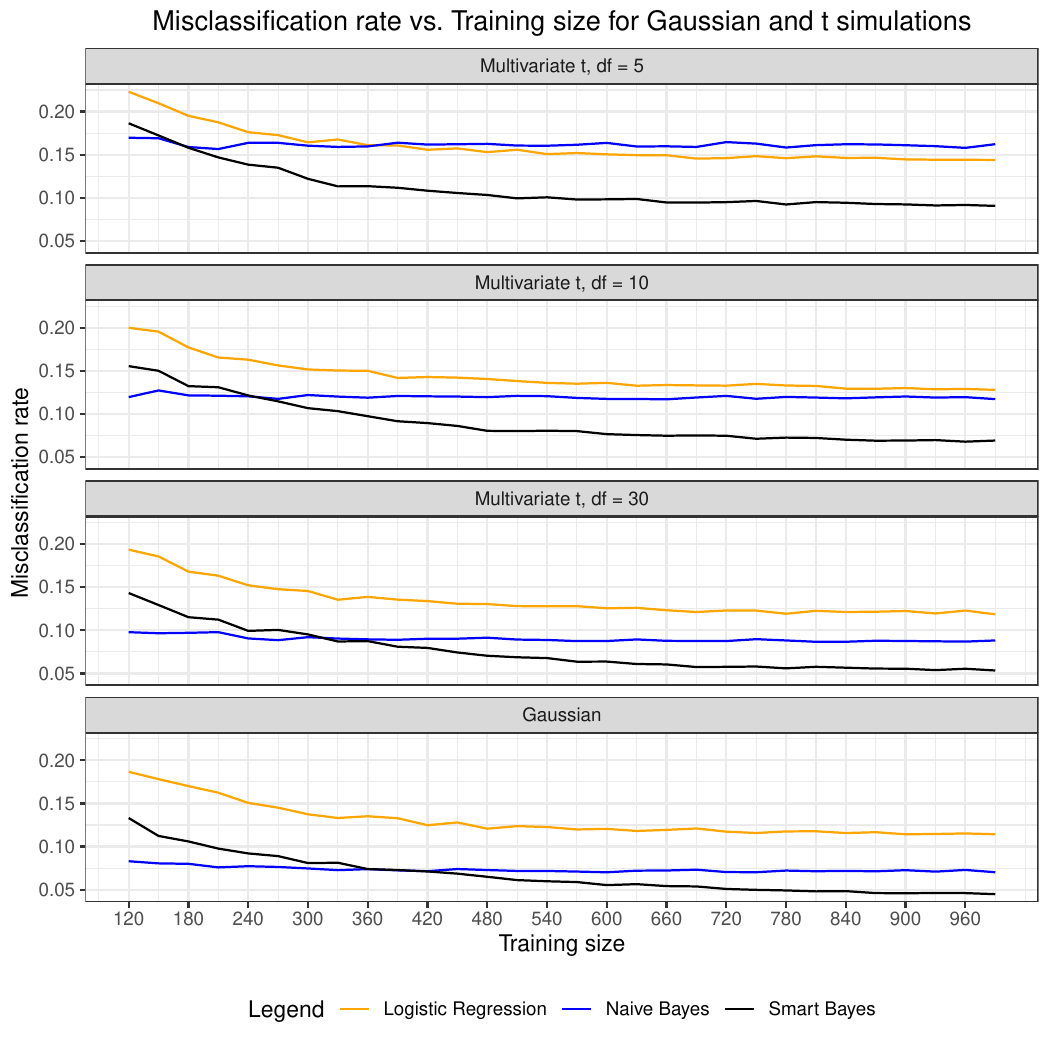}
\caption{Average misclassification rate over 100 replications at different training sizes for the multivariate $t$ with 5, 10, and 30 degrees of freedom and the Gaussian. Error rates were computed at every 30 observations. These simulations used $p = 32$ covariates, where the mean vector $\mathbf{\mu}$ and the covariances $\mathbf{\Sigma}$ used in the data generation were the sample means and covariances for the two classes in the ionosphere dataset.}
\label{fig:sim-gauss-t-ion}
\end{figure}

In this simulation, we consider four data-generating distributions: a multivariate $t$ with degrees of freedom $df=5$, $df=10$, and $df=30$, as well as a multivariate Gaussian distribution. For each training size $m$, we generate $m$ observations from Class 0 and $m$ observations from Class 1. We then randomly sample half of all the observations and assign them to the training set and assign the remainder to the test set. The means and covariances for the classes in this simulation are the sample means and covariances from the two classes of the ionosphere dataset in Section~\ref{sect-realdata}. Figure~\ref{fig:sim-gauss-t-ion} reports the average misclassification rate over 100 replications for Naive Bayes, Smart Bayes, and logistic regression.

In each of these simulations, Smart Bayes has the lowest misclassification rate as the training size grows large. Smart Bayes always outperforms logistic regression and outperforms Naive Bayes at large sample sizes; however, when the data generating distributions have heavier tails (fewer degrees of freedom), Smart Bayes overtakes Naive Bayes even at smaller sample sizes. Naive Bayes appears to have a lower asymptotic error rate than logistic regression as the degrees of freedom increase.

\subsection{Real data experiments}
\label{sect-realdata}
We evaluate the performance of the proposed Smart Bayes classifier versus Naive Bayes and logistic regression on seven real datasets from the UCI Machine Learning Repository~\cite{ucirepo}. All datasets except for the abalone dataset were used in Ng and Jordan~\cite{ng2002discriminative}. We reproduce their comparisons while adding the Smart Bayes classifier using the spline-based density-ratio estimation method. 

As done by Ng and Jordan~\cite{ng2002discriminative}, we evaluate the performance of these classifiers at different training sizes. At each training size $m$, we randomly select $m$ observations as training samples and assign the rest of the observations in the dataset to the test sample. We conduct 200 replications at each training size and report the average misclassification rate on out-of-sample data. These are plotted in the figures below. The general shapes of the plots below match those of the same figures in~\cite{ng2002discriminative}, but there are minor differences. Although we attempted to choose the same variables as Ng and Jordan, we may have omitted or included variables that they did not omit or include, since their description only states that non-continuous variables were excluded from the analysis. It is also likely that their implementations of logistic regression and Naive Bayes differ from those used here. Experiments were conducted on an AWS EC2 instance of size \texttt{t2.xlarge} using the \texttt{clustermq} package~\cite{clustermq} in version 4.3.2 of R~\cite{Rcitation}.

\begin{figure}[H]
\centering
\includegraphics[width=.8\textwidth]{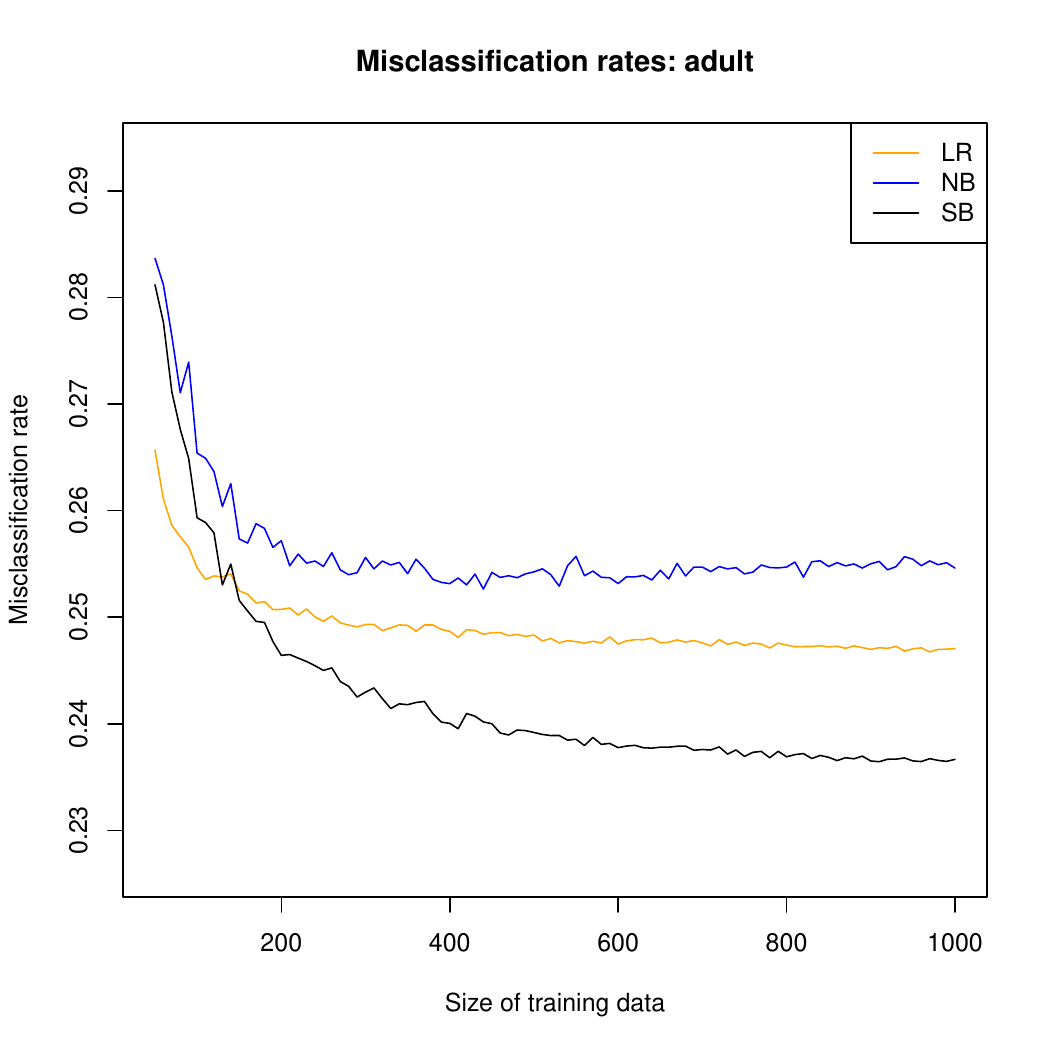}
\caption{Average misclassification rate over 200 replications at different training sizes for the adult dataset. Non-continuous variables were removed from the classification.}
\label{fig:adult}
\end{figure}

\begin{figure}[H]
\centering
\includegraphics[width=.8\textwidth]{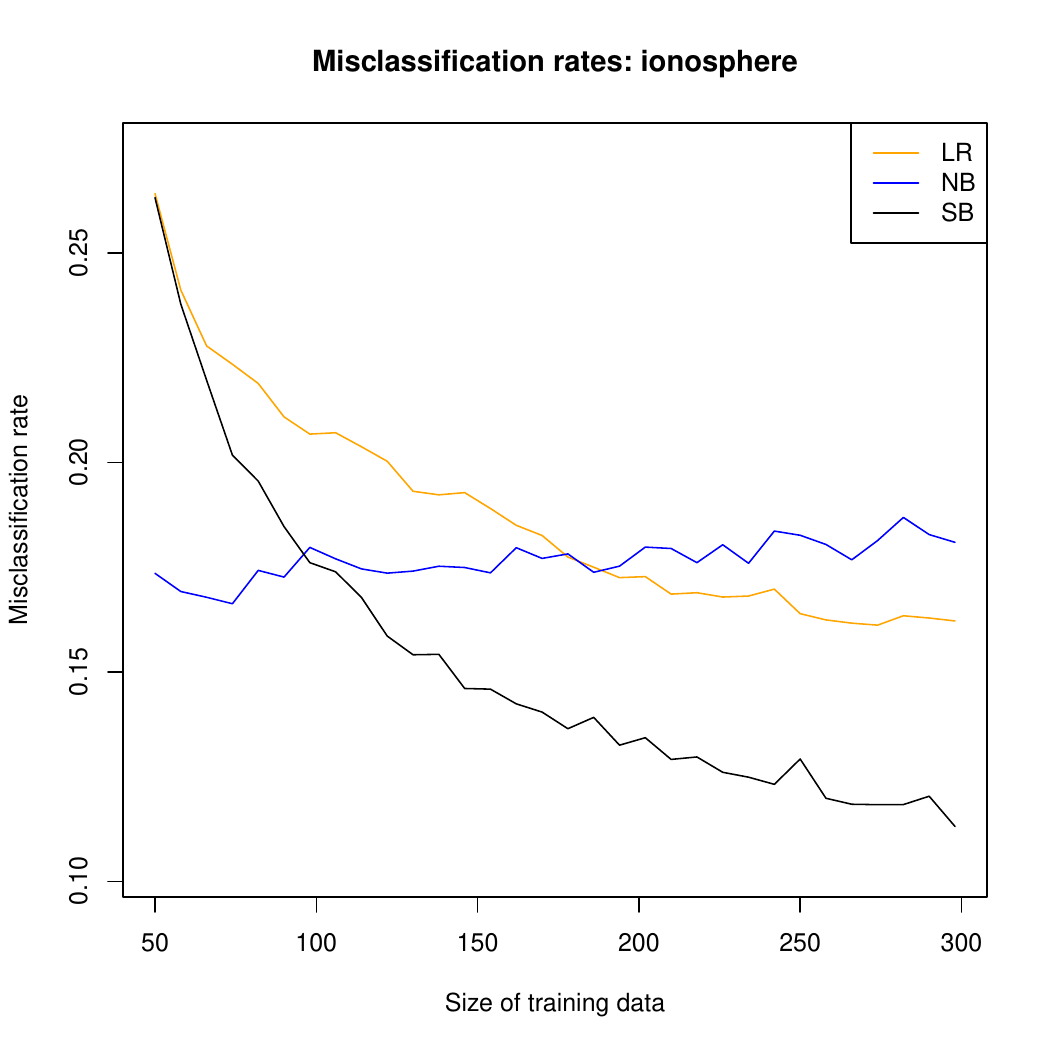}
\caption{Average misclassification rate over 200 replications at different training sizes for the ionosphere dataset. Non-continuous variables were removed from the classification.}
\label{fig:ionosphere}
\end{figure}

\begin{figure}[H]
\centering
\includegraphics[width=.8\textwidth]{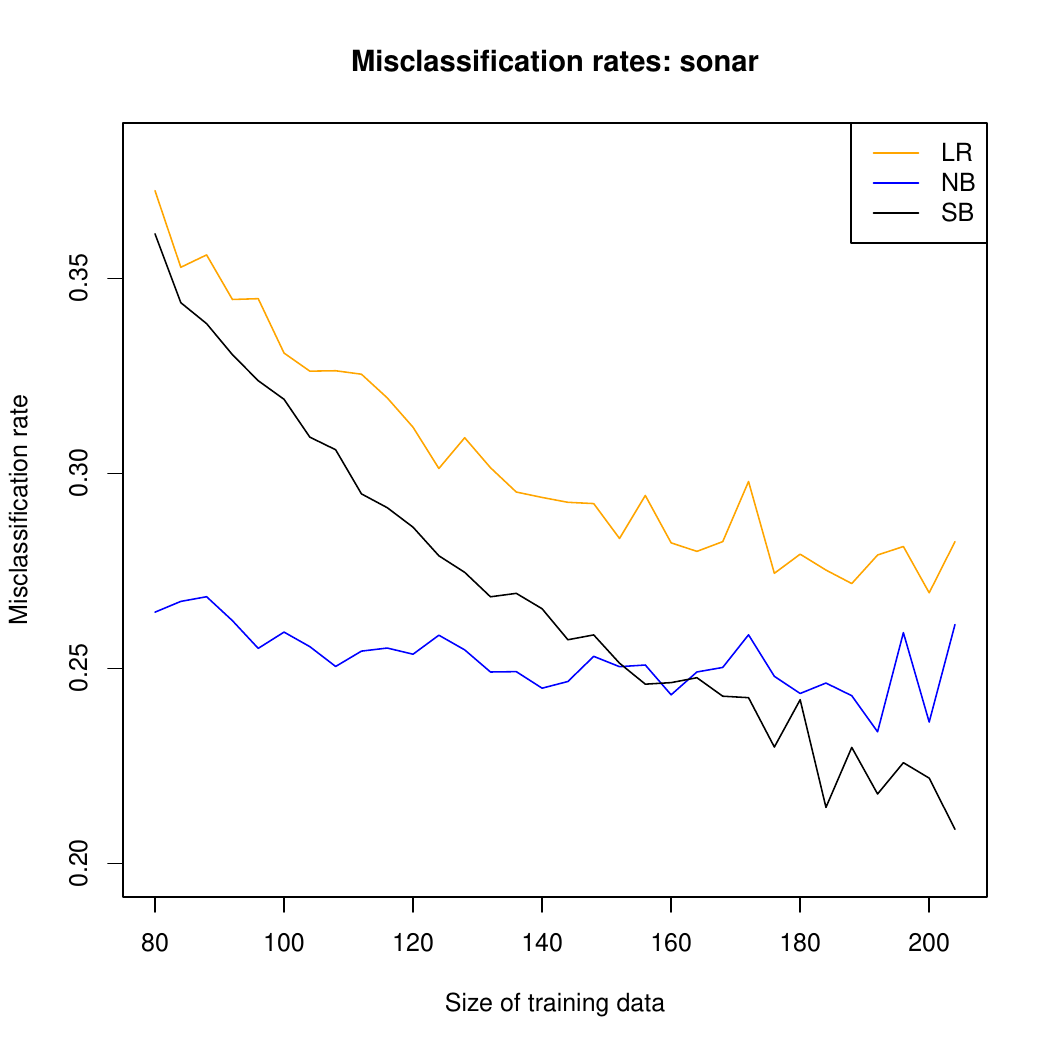}
\caption{Average misclassification rate over 200 replications at different training sizes for the Sonar dataset. Non-continuous variables were removed from the classification.}
\label{fig:sonar}
\end{figure}

In the experiments, there are several instances where the Smart Bayes classifier outperforms both Naive Bayes and logistic regression. In particular, the adult dataset in Figure~\ref{fig:adult}, the ionosphere dataset in Figure~\ref{fig:ionosphere}, and the sonar dataset in Figure~\ref{fig:sonar} all demonstrate that the Smart Bayes classifier can yield lower misclassification rates than the other methods. These differences generally arise at larger training sizes, mimicking logistic regression's relationship with Naive Bayes as described by~\cite{ng2002discriminative}. The Smart Bayes classifier also relies on an ability to estimate density ratios, which the Naive Bayes classifier requires and logistic regression does not. These two properties reinforce Smart Bayes's need for sufficient training observations.



\begin{figure}[H]
\centering
\includegraphics[width=.8\textwidth]{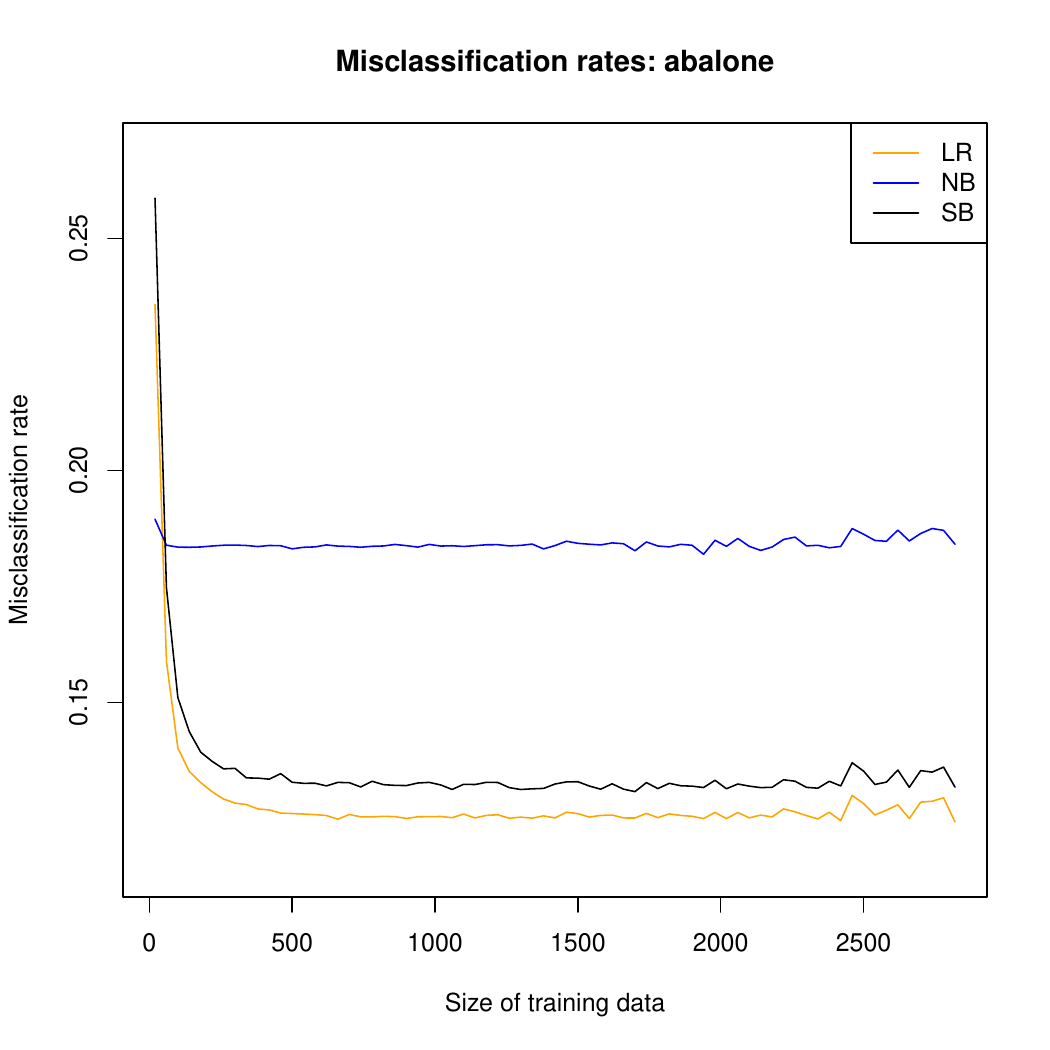}
\caption{Average misclassification rate over 200 replications at different training sizes for the abalone dataset. Observations corresponding to the middle two quartiles of the response variable were removed from the dataset, so classification took place on data that were above the 3rd quartile (Class 1) or below the 1st quartile (Class 0).}
\label{fig:abalone}
\end{figure}

\begin{figure}[H]
\centering
\includegraphics[width=.8\textwidth]{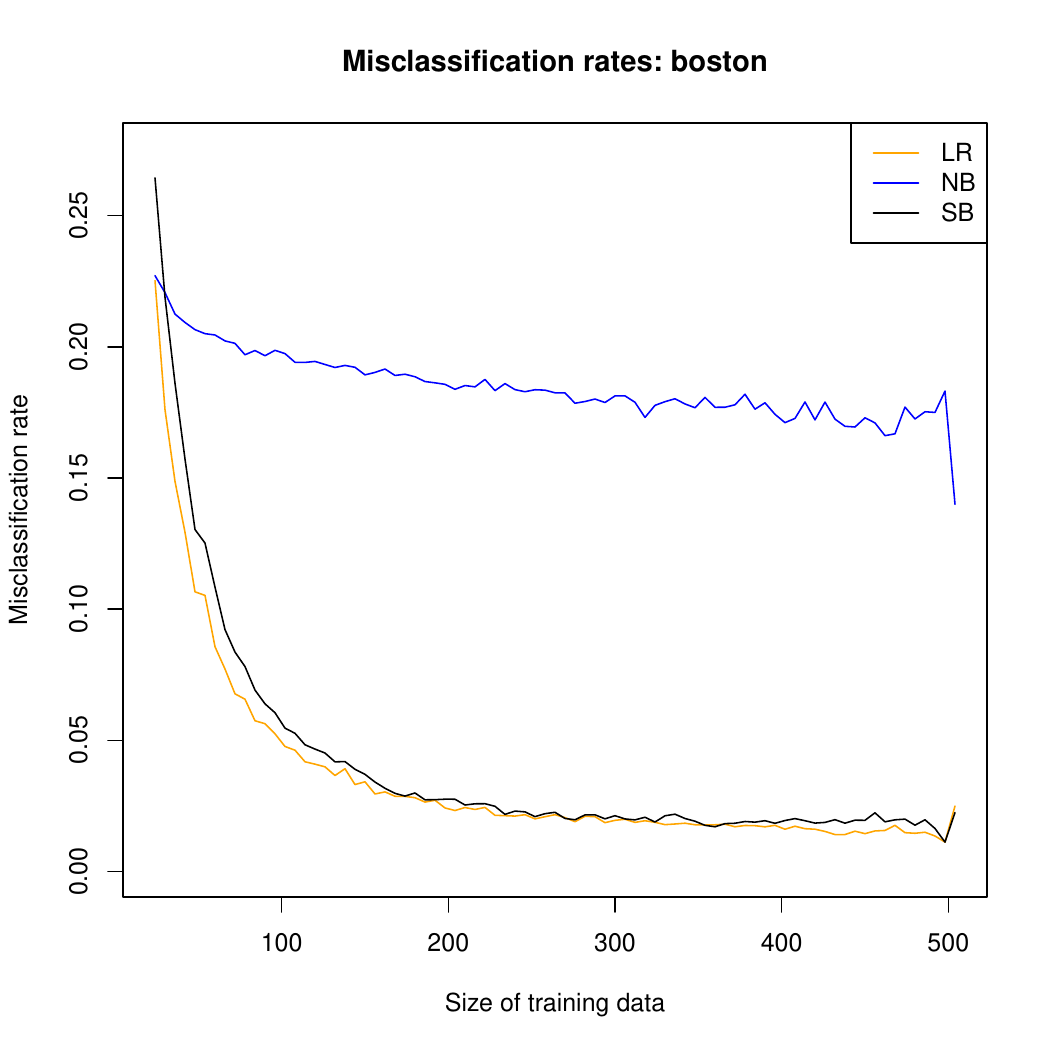}
\caption{Average misclassification rate over 200 replications at different training sizes for the Boston housing dataset. The response variable was whether the median value of the home exceeded the median value of all homes. Non-continuous variables were removed from the classification. This dataset is named \texttt{Boston} and is available via the MASS library~\cite{mass}.}
\label{fig:bos}
\end{figure}

\begin{figure}[H]
\centering
\includegraphics[width=.8\textwidth]{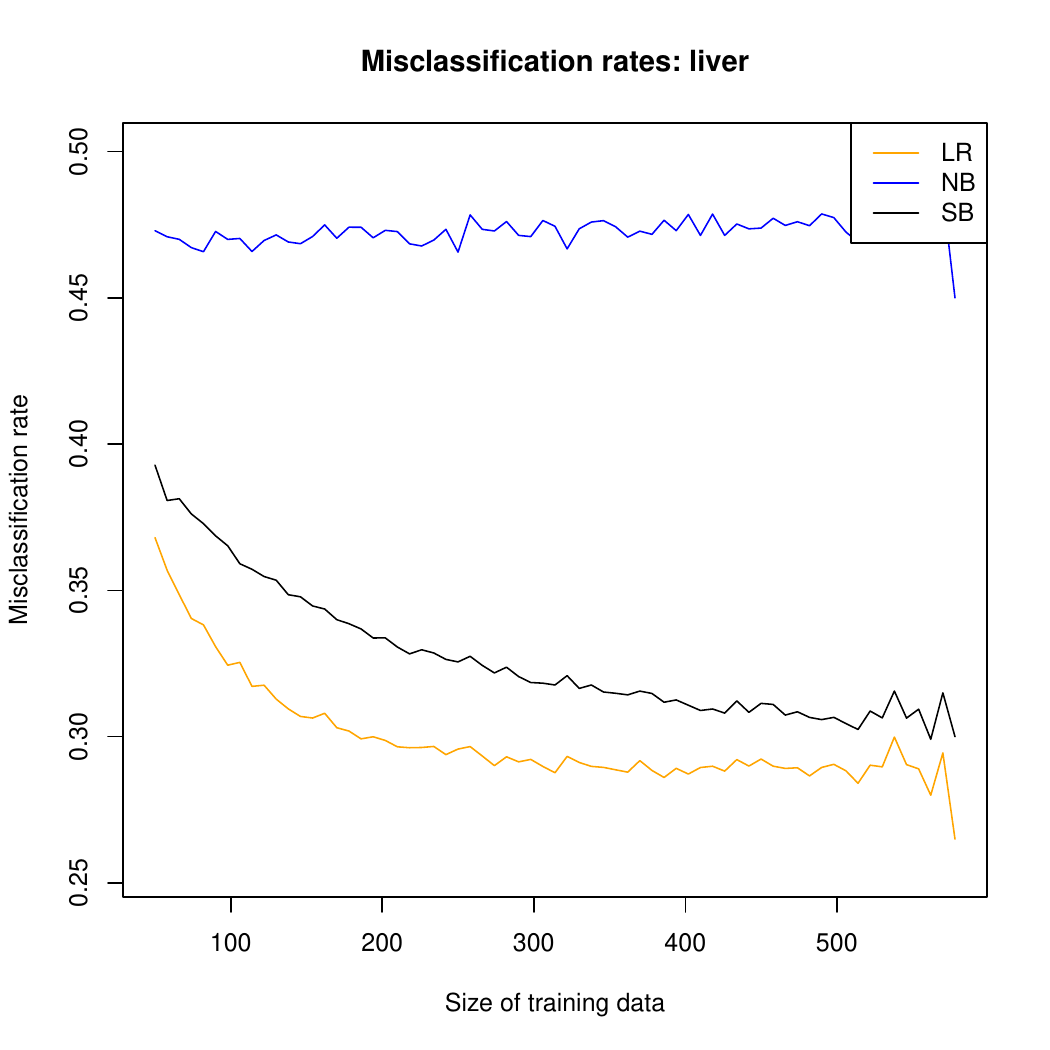}
\caption{Average misclassification rate over 200 replications at different training sizes for the liver dataset. Non-continuous variables were removed from the classification.}
\label{fig:liver}
\end{figure}

\begin{figure}[H]
\centering
\includegraphics[width=.8\textwidth]{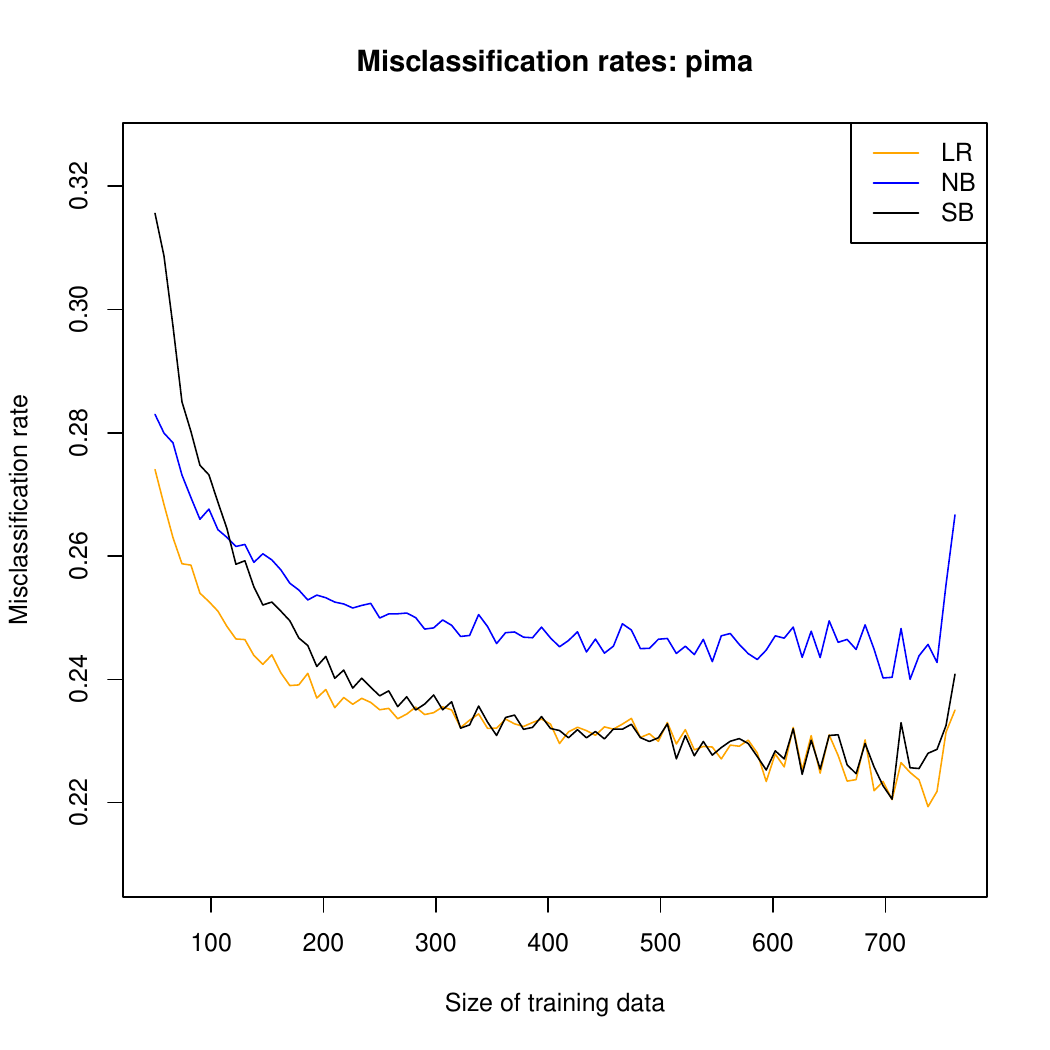}
\caption{Average misclassification rate over 200 replications at different training sizes for the Pima dataset. Non-continuous variables were removed from the classification.}
\label{fig:pima}
\end{figure}

\section{Discussion}
\label{sect-discussion}
We proposed a hybrid classifier, Smart Bayes, that uses density ratios as features within a logistic regression framework. Smart Bayes generalizes Naive Bayes by allowing flexible weighting of density ratios. To build the classifier, we developed a spline-based method for estimating log density ratios in Section~\ref{sect-density}.

We compared the performance of the Smart Bayes classifier to both Naive Bayes and logistic regression in simulation experiments in Section~\ref{sect-subsim} and in real data experiments in Section~\ref{sect-realdata}. The simulations show cases where Smart Bayes outperforms logistic regression and Naive Bayes as the training size increases. The simulations also demonstrate that there are cases where Naive Bayes appears to have a lower asymptotic error rate than logistic regression, as discussed in~\cite{xue2008comment}.

The empirical experiments, which use some of the same datasets and are presented in the style of~\cite{ng2002discriminative}, also show that Smart Bayes can outperform its competitor methods. We do not present a theoretical explanation for when Smart Bayes outperforms the other two methods. As noted in~\cite{xue2008comment}, there is no general criterion for deciding whether to use a discriminative or generative classifier. However, this work demonstrates that there may benefits to blending the two approaches.

There are several promising directions for future research. Smart Bayes may be enhanced by incorporating additional terms, such as higher-order interactions between generative features, and by using regularization methods to select among them. While this work focused on features derived from likelihood ratios of univariate marginal densities, extending the approach to multivariate marginal densities, guided by prior knowledge or graphical structure, may yield further improvements and produce classifiers closer to the optimal Bayes rule. Theoretical work that delineates when and why Smart Bayes outperforms Naive Bayes and logistic regression would also be worthwhile.

Another direction is to integrate these generative features into other machine learning models, such as support vector machines, transformers, and other deep learning architectures to assess whether they offer similar gains in classification performance. Propensity score matching is also an area where Smart Bayes could be useful, since its potential to approach the optimal Bayes classifier may yield better propensity scores than logistic regression.

\bibliographystyle{plain}
\bibliography{references}
\end{document}